\documentclass[8pt, twocolumn, a4paper]{extarticle}
\usepackage[top=2cm, left=2cm, right=2cm, bottom=2cm]{geometry}
\usepackage{float}
\usepackage{balance} 
\usepackage{graphicx}
\usepackage{subfigure}
\usepackage{xfrac}
\usepackage{bm}
\usepackage{amsmath}
\usepackage{amsfonts}
\usepackage{mathtools}
\usepackage{amssymb}
\usepackage{amsthm}
\usepackage{adjustbox}
\usepackage[font=small,labelfont=it]{caption}
\usepackage{wrapfig}
\usepackage{enumitem}
\usepackage{doi}
\usepackage{hyperref}
\usepackage[small]{titlesec}
\usepackage[labelfont=bf]{caption}
\usepackage{subcaption}
\usepackage{xcolor}
\usepackage{booktabs}
\usepackage{multicol}
\usepackage{multirow}
\usepackage{stfloats}

\newtheorem{theorem}{Theorem}

\usepackage{algorithm}
\usepackage{algorithmic}

\usepackage[numbers]{natbib}
\title{Neural Mean-Field Games: Extending Mean-Field Game Theory with Neural Stochastic Differential Equations}

\author{
  {\normalsize \textbf{Anna C.M. Thöni}}\\
  {\normalsize Donders Centre for Cognition}\\ 
  {\normalsize Radboud University}\\
  {\normalsize Nijmegen, The Netherlands}\\
  {\normalsize \texttt{chiara.thoeni@donders.ru.nl}}
  \and
  {\normalsize\textbf{Yoram Bachrach}} \\
  {\normalsize Meta FAIR}\\
  {\normalsize London, United Kingdom}\\
  {\normalsize \texttt{yorambac@meta.com}}
  \and 
  {\normalsize\textbf{Tal Kachman}}\\
  {\normalsize Donders Centre for Cognition}\\
  {\normalsize Radboud University}\\
  {\normalsize Nijmegen, The Netherlands}\\
  {\normalsize \texttt{tal.kachman@donders.ru.nl}}
}
\date{}

\begin{document}

\maketitle 
\begin{abstract}
Mean-field game theory relies on approximating games that are intractable to model due to a very large to infinite population of players. While these kinds of games can be solved analytically via the associated system of partial derivatives, this approach is not model-free, can lead to the loss of the existence or uniqueness of solutions, and may suffer from modelling bias. To reduce the dependency between the model and the game, we introduce neural mean-field games: a combination of mean-field game theory and deep learning in the form of neural stochastic differential equations. The resulting model is data-driven, lightweight, and can learn extensive strategic interactions that are hard to capture using mean-field theory alone. In addition, the model is based on automatic differentiation, making it more robust and objective than approaches based on finite differences. We highlight the efficiency and flexibility of our approach by solving two mean-field games that vary in their complexity, observability, and the presence of noise. Lastly, we illustrate the model's robustness by simulating viral dynamics based on real-world data. Here, we demonstrate that the model's ability to learn from real-world data helps to accurately model the evolution of an epidemic outbreak. Using these results, we show that the model is flexible, generalizable, and requires few observations to learn the distribution underlying the data.
\end{abstract}

\section{Introduction}
\label{introduction}

Game theory investigates strategic interactions between self-interested agents that inhabit the same environment and are affected by each other's actions. Such analysis has applications in economics~\cite{gameTheoryEconomics1992, decisionMakingMarkets2010, achdouMacroEconomics}, crowd motions \cite{achdou2019mean, aurell2019} and evolutionary biology \cite{evolutionaryBiology2021}. Some of these applications may consider an extremely large, or in the limit, infinite population. As it is intractable to model all interactions between such a large number of agents, these types of games are approximated using mean-field game (MFG) theory \cite{LasryLionsMFG2007, HuangMFG2006}. MFG theory replaces the agents with their mean field, effectively summarizing their interactions by a distribution \cite{Guéant2011}. From a mathematical perspective, this distribution is described with two differential equations that are coupled through the agents' cost function: the first, a Hamilton-Jacobi-Bellman equation, guides the optimization strategy of the average agent, whereas the second, a Fokker-Planck equation, describes the temporal evolution of the population's overall dynamics.

Even though each agent influences the population's overall dynamics by minimizing their own expected costs, the infinitesimal agents are too small to independently influence the population. Instead, all agents are expected to jointly converge towards an optimal control that minimizes their individual costs. This state, which can be considered the game's Nash equilibrium, can be found by numerically solving the partial or stochastic differential equations that accompany the game \cite{CardaliaguetNotes}. This can, for example, be done by numerically solving the differential equations using finite differences \cite{ Briceno-Arias2019, MFGnumericalMethods}. Even so, solving a system of differential equations involving high-dimensional agent states is challenged by the curse of dimensionality \cite{learningInMFGSurvey, bellman1957markovian}. In addition, the discretization introduces an error that is subject to the chosen method and may ultimately lead to the loss of the existence or uniqueness of the solutions \cite{RL-basedNNs}.
Instead, the solutions to the mean-field game can be approximated using artificial neural networks \cite{TongLinNN}, in the form of recurrent architectures \cite{CarmonaNNapprox, CarmonaNNapprox2}, deep policy iteration \cite{ASSOULI2024128923}, deep Galerkin methods \cite{ASSOULI_DG}, or actor-critic-based approaches \cite{ActorCriticApproximation, RL-basedNNs, ZHOU20208112}. Still, all approaches mentioned above require full knowledge of the game of interest. This dependency limits the models' applications and may introduce a modelling bias. Reinforcement learning (RL)-based solutions surmount this requirement \cite{Lazaridis2020}, for example, by building upon the Fictitious Play algorithm instead of following a numerically exact approach \cite{Subramanian2020PartiallyOM, DRL, MFG_DRL2021, anahtarciQLMFG2023, RLMishra, Lauriere2022}. Because the Fictitious Play algorithm is independent of the modelled system, the RL-based algorithms are commonly referred to as ``model-free" \cite{learningInMFGSurvey}.
\begin{figure}[b]
    \centering
    \includegraphics[width=0.5\textwidth]{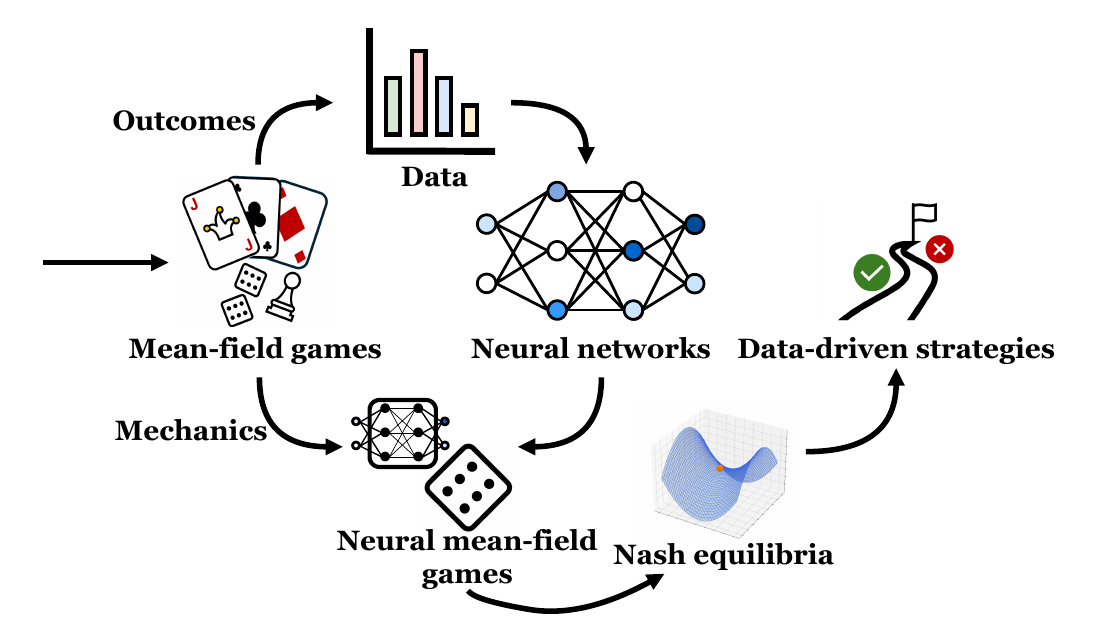}
    \caption{Overview of modelling MFGs with neural SDEs. The resulting neural mean-field game combines the MFG mechanics with the neural network output to create more informed strategies.}
    \label{fig:overview}
\end{figure}
To retain the numerical exactness of the differential equation-based solutions without requiring full knowledge of the game of interest, we propose to analyze MFGs using neural stochastic differential equations (SDEs; \cite{Tzen2019, Liu2019}). Neural SDEs form the stochastic extension of neural ordinary differential equations (ODEs), a family of neural networks introduced by \citet{chen2018neural}. To model the MFG dynamics as a neural SDE, we complement the existing system of coupled differential equations with a neural network. The resulting model, shown in Figure \ref{fig:overview}, is data-driven and may therefore alleviate the modelling bias associated with the predefined MFG dynamics. Furthermore, the model's ability to recover structure from data allows it to reason beyond the predefined differential equations. Lastly, the neural SDEs are solved using automatic differentiation, which makes them more robust than approaches driven by classical numerical methods such as finite differences and finite elements \cite{SU1997}. This property, in combination with the model's data-driven nature, is expected to prevent the loss of the existence of solutions that affect the original differential equation-based approaches. We demonstrate the flexibility of the neural MFGs by modelling the dynamics of COVID-19 infections. Dynamical systems for epidemics include complex interactions \cite{Jia2025}, are highly nonlinear and depend on various features, including population parameters, restrictive measures, and viral characteristics \cite{Marinov2022}. Because it can be challenging to quantify the effects of such features, their inclusion is often beyond the scope of classical mathematical models \cite{Marinov2022, Gaskin2023}. To provide insight into the complex mechanics behind epidemic systems, we extend the classical models by learning viral mechanics from a large range of epidemic characteristics.\clearpage
\noindent
\textbf{Contributions} The contributions of this paper are fourfold: (1) We introduce neural MFGs, providing the theoretical framework for combining MFG theory with neural SDEs, (2) we reformulate examples from classical game theory to their MFG counterparts, taking the continuum limit on the number of agents by leveraging the structure of the neural SDEs, and (3) we test the neural MFGs on the two games with increasing complexity to give a proof of concept of the model's ability to construct new, data-driven Nash equilibria by empirically highlighting emerging strategies. Lastly, (4) we apply the neural MFGs on a real-world example, by augmenting a compartmental model of the COVID-19 outbreak in Japan and incorporating auxiliary observations. \\[3mm]
\textbf{Data and Code Availability} The work uses data obtained from \citet{MOHJapan} and \citet{Edouard2020}. The code associated with the model implementation and analyses is available on GitHub: \url{https://github.com/KachmanLab/NeuralMFG}.
\section{Background}
In this section, we review MFG theory and the neural ODEs and SDEs introduced in \cite{Liu2019, chen2018neural}. MFG theory, which has largely been shaped by a series of contributions by Lasry and Lions \cite{Lasry2006a, Lasry2006b, LasryLionsMFG2007}, proposes an approximation of the Nash equilibria for stochastic differential games with a large number of players \cite{Carmona2013}. As mentioned before, MFGs can be described using both partial differential equations (PDEs) and concepts from RL. This section considers the PDE perspective, which allows for an analytical and tractable formulation of MFGs. In addition, treating the formulation based on differential equations makes it easier to unite MFGs with the neural SDEs discussed later. 

\subsection{Modelling Games with an Infinite Number of Players}\label{sec: MFG}
MFGs model the dynamics of the infinitesimal players and the full population by jointly solving the Hamilton-Jacobi-Bellman (HJB) equation and the Fokker-Planck (FP) equation. To illustrate this coupling, we consider a stochastic differential game with $N$ players. Each player $i \in [1, \dots, N]$ has a state at time $t \in [t_0, T]$, denoted with $x_t^i \in \mathbb{R}^d$. The state of agent $i$ evolves over time according to $\alpha_t^i$: the player's control, or game-playing strategy, at time $t$. The control is a function that takes values within the set of admissible controls $\mathcal{A}$. We assume that the dynamics of a single player are given by a (general) Itô process \cite[p. 21]{Oksendal2013} of the form
\begin{equation}
    d x^i_t = b\left(t, x_t^i, \alpha_t^i, m(t)\right)\,dt +  \sigma(t, x_t^i, \alpha_t^i, m(t))\,dB_t
    \label{eq: dynamics}
\end{equation}
where the deterministic and stochastic terms $b$ and $\sigma$ respectively describe the player's drift and diffusion, and $B$ is an Wiener process (Brownian motion). Naturally, the individual players may base their strategy on other players' actions. This dependency is accounted for with the inclusion of $m(t) = \frac{1}{N}\sum^N_{i=1} \delta_{x_t^i}$, the empirical distribution representing the mean-field of all other players' states $x_t = (x_t^1, \dots, x_t^N)$ \cite{CardaliaguetNotes, Carmona2013}. In this formulation, $\delta_x$ denotes the Dirac measure at $x$. 

\subsection{Mean-Field Game Principles} An MFG can be seen as an optimization problem, where each player chooses a strategy that minimizes their expected cost over the period $[t_0, T]$. This agent-specific cost, $J^i$, is given by the quantity
\begin{equation}
    J^i = \mathbb{E}\left[\int_{t_0}^T \frac{1}{2} L(x_s^i, m(s), \alpha_s^i)ds + G(x_T^i, m(T))\right].
    \label{eq: cost}
\end{equation}
Following the fundamentals of dynamic programming, $J$ combines the evolution of the player's cost $L$ through $s$ with a terminal cost $G$ at time $T$ \cite{CardaliaguetNotes}. Just like the agents' dynamics, the cost function links the agents' individual states $x$ to the mean-field $m$. Assuming that agent $i$ is a representative agent in the system, it is implied that all agents aim to minimize the cost function. Therefore, MFGs are commonly analyzed with the generalized notion of the Nash equilibrium, which is reached for a family of controls ($\bar{\alpha} \in \mathcal{A})$ when 
\begin{equation}
    \forall i, j \in [1, \dots, N], \; J^i\left(\bar{\alpha}^i, \bar{\alpha}^j_{j \neq i}\right) \leq J^i\left(\alpha^i, \bar{\alpha}^j_{j\neq i}\right)
\end{equation}
holds for all non-Nash strategies $\alpha^i$ \cite{peters2008game}. That is, the Nash equilibrium is a fixed point in the space of the flow in $m$ \cite{carmona2018probabilistic}, which implies that no agent benefits from changing their strategy if a Nash equilibrium is reached. Computing the Nash equilibria for many agents involves a large set of controls and can be a lot of work. Fortunately, MFG theory simplifies this problem with the assumptions of symmetry and homogeneity \cite{Guéant2011}, facilitating the calculation of Nash equilibria when $N \to \infty$. In short, these assumptions imply that all players $i$ are statistically identical and can be described with the same drift $b$ and diffusion $\sigma$ \cite{Carmona2013}. We refer the reader to \citet{CardaliaguetNotes} for a detailed account of this simplification. 

To facilitate the approximation of the Nash equilibria when $N\to \infty$, \citet{LasryLionsMFG2007} formalize MFGs as a system of PDEs, as shown in Equation \ref{eq: MFG}.  The system consists of (a) a HJB equation providing the necessary and sufficient conditions for the optimality of $\alpha$ w.r.t.\@ $J$ and (b) a FP equation describing the evolution of the mean-field according to the dynamics presented in Equation \ref{eq: dynamics}.\\[2mm]
\resizebox{\linewidth}{!}{
\begin{minipage}{1.3\linewidth}
\begin{equation}
\begin{cases}
       \text{(a)}\quad  - \partial_tu - \nu \Delta u + H(x_t^i, m(t), Du) = 0 & \text{in $\mathbb{R}^d \times (0, T)$}\\
       \text{(b)}\quad  \partial_tm(t) - \nu \Delta m(t) -\text{div}\left(D_pH\left(x_t^i, m(t), Du\right)m(t)\right) = 0 & \text{in $\mathbb{R}^d \times (0, T)$}\\
    \text{(c)} \quad m_0 = m(0),\; u(x, T) = G(x_T^i, m(T))
    \end{cases}
\label{eq: MFG}
\end{equation}
\end{minipage}
}
\newline
The coupling of the HJB and FP equations is facilitated via the unknowns $u$ and $m$. In particular, the first equation goes backwards in time and describes the value function of a single agent, $u$.\footnote{$u$ can be considered the supremum of the optimization problem described by the functional $J$, that is, $u(x, t) = \underset{\alpha \in \mathcal{A}}\inf \; J(\alpha)$ \cite{Achdou2020, Cardaliaguet2021}.} The second goes forward in time and describes the change in the mean-field density $m(t)$. The equation is parameterized with the nonlinear Hamiltonian $H$, which is the Fenchel conjugate to $L$ w.r.t.\@ the mean-field probability $p$. In addition, $D$ denotes a differential operator and $\nu$ is a nonnegative parameter. Both equations are subject to two structure conditions \cite{LasryLionsMFG2007}: Firstly, $H(x, m, p)$ is convex with respect to $p$.\footnote{This is required to satisfy the optimality condition provided by the HJB equation \cite{LasryLionsMFG2007}.} Secondly, $m$ is a probability measure with the initial state $m(0)$ (Equation \ref{eq: MFG}c). In addition to the structure conditions, the terminal condition $u(x, T) = G(x_T^i, m(T))$ relates Equation \ref{eq: MFG} back to the cost presented in Equation \ref{eq: cost} \cite{LasryLionsMFG2007, CardaliaguetNotes}. 

\subsection{Neural Differential Equations}\label{sec: nODE}
Neural ODEs, and neural SDEs in their extension, are a neural network-based method introduced by \citet{chen2018neural}. To introduce the method, we consider a simple ODE $h$ describing the drift $b$ of $x$ over time $t$.
\begin{equation}
    h(t, x(t)) = b(t, x) \, dt. 
\end{equation}

To learn the development of the dynamics of $x$, we can solve the initial value problem (IVP) consisting of the initial state $x(t_0) = x_0$ and the ODE describing their dynamics $\sfrac{dx}{dt}=h(t, x(t))$. The solution to this problem is commonly calculated using discretization, transforming the continuous function $h$ into a counterpart that consists of small steps \cite{Calculus2013}. In this case, $x(t)$ is updated step-by-step by adding its identity to $h(t, x(t))$, the output of the ODE. This process is illustrated in Equation \ref{eq: vanillaODE}, where $\Delta t$ denotes the discretized stepsize. 
\begin{equation}    
x(t + \Delta t) = x(t) + h(t, x(t))
\label{eq: vanillaODE}
\end{equation}

\citet{chen2018neural} highlight the similarity between the discretization of continuous functions and the residual motif $y = F(x) + x$ from ResNet architectures \cite{he2015deep}. They compare the addition of the identity $x(t)$ to the residual skip connections, relating the discretization steps to the repeated application of the residual blocks $F$. This analogy lays the foundation for neural ODEs: a class of differential equations expressed as a neural network. Thus, instead of discretizing a concrete mathematical function $h$, the neural ODE repeatedly applies the data-driven output of a neural network $f_\theta$. The discretized step size, $\Delta t$ is optimized by the differential equation solver and may vary throughout the solving process. As a consequence, a neural ODE can be seen as a neural network with a flexible number of layers. The output of the solver can be regarded as a latent trajectory over time $t \in [t_0, T]$, where each output $\bm{z}_{t_0}, \dots, \bm{z}_{T}$ represents the latent state \cite{chen2018neural}. Because the neural ODEs are trained using reverse automatic differentiation and the adjoint sensitivity method \cite{pontryagin2018mathematical}, the user has explicit control of the model's numerical error \cite{chen2018neural}. 

\subsubsection{Modelling Real-World Stochastic Processes} Instead of merely modelling a deterministic function, neural SDEs extend the neural ODEs to deal with the presence of stochasticity and randomness underlying the process \cite{Tzen2019, Jia2019, Peluchetti2020}. In a neural SDE (see Equation \ref{eq: neuralSDE} and \cite{Li2020}), both the drift $\mu_\theta$ and diffusion $\sigma_\theta$ are expressed as a neural network. Thus, instead of learning a single path, a neural SDE learns the \emph{distribution} over path-valued random variables. The existence and uniqueness of a strong solution for $y$ is guaranteed as long as (1) $\mu_\theta$ and $\sigma_\theta$ are Lipschitz, which typically holds for neural networks, and (2) $\mathbb{E}[y(0)^2] < \infty$ \cite{kidger2021}. In Equation \ref{eq: neuralSDE}, which shows the neural SDE dynamics, $\theta$ is used to refer to the neural network parameters. 

\begin{equation}
    dy(t) = \mu_\theta(t, y(t))\, dt + \sigma_\theta(t, y(t))\, dB_t
    \label{eq: neuralSDE}
\end{equation}

Because both the neural ODEs and SDEs are solely specified in terms of neural networks, the methods are data-driven. All the while, dynamical systems may be described using domain knowledge based on scientific models or physical laws. \citet{rackauckas2020universal} unify the above with the introduction of universal differential equations (UDEs), combining the output of $f$ with that of a concrete ODE $h$: $\sfrac{dx}{dt}= h(t, x(t)) + f(x)$. In this case, $f$ can be trained to yield insights that go beyond the predefined ODE $h$. A stochastic UDE variant has already been used to solve an HJB equation, highlighting the model's adaptivity, efficiency, and flexibility towards stiffness \cite{rackauckas2020universal}.  

\section{Neural Mean-Field Games}
We introduce neural Mean-field games, a new perspective on UDEs that combines existing MFG specifications with neural SDEs. To do so, we include the neural SDE at the agent level. This way, the neural network output may provide additional insights drawn from the mean-field, past game states, or observational data. Because these insights directly inform the agents' value function $u$, the agents are expected to make better-informed decisions. In addition, this approach leverages the setup of the neural SDE, which allows for taking the number of agents or turns to the continuum limit. 

\subsection{Data-Driven Dynamics}
The addition of neural networks to the MFG dynamics has implications for the formulation. In particular, we update the description of the single-player dynamics (Equation \ref{eq: dynamics}) by adding the neural drift from Equation \ref{eq: neuralSDE} and modelling the diffusion in a fully data-driven manner. This results in the general formulation of the neural MFG dynamics, presented in equation \ref{eq: neuralMFG dynamics}.
\\
\resizebox{\linewidth}{!}{
\begin{minipage}{1.2\linewidth}
\begin{equation}
d x^i_t = \left[b\left(t, x_t^i, \alpha_t^i, m(t)\right) + \mu_\theta(t, x_t^i, a_t^i, m(t))\right]\,dt +  \sigma_\theta(t, x_t^i, \alpha_t^i, m(t))\,dB_t
    \label{eq: neuralMFG dynamics}
\end{equation}
\end{minipage}
}
We make the neural MFG data-driven by incorporating a loss term that minimizes the discrepancy between the players' states and observations. In this case, we extend the cost function $J^i$ with a loss term comparing the predicted distribution at time to the observed label. This loss term can be included in the running costs $L$, or in the terminal costs $G$, depending on the MFG of interest. Thus, in contrast to a single MFG, a neural MFG minimizes the sum of two losses: one originating from the game description and one minimizing the discrepancy between the model predictions and observed data. As the form of the combined loss depends on the game at hand, we will provide an instantiation and interpretation of the loss in the following sections. Because of the symmetry assumption and the joint minimization of the players' loss, the game's Nash equilibria now depend on $b$, $\mu_\theta$, and $\sigma_\theta$ instead of only on $b$ and $\sigma$. As a result, the Nash equilibria become data-driven. 

\paragraph{A Note on the Existence and Uniqueness of Solutions}\label{sec: existence}
Naturally, the inclusion of the neural networks has implications for the existence and uniqueness of the solutions to the MFG. However, we show that the formalism behind the existence and uniqueness of MFG solutions can be extended to the neural MFG case. We consider two types of MFGs: games with a continuous state space, as introduced by \cite{LasryLionsMFG2007} and \citet{HuangMFG2006}, as well as games with a discrete state space, as presented by \citet{Doncel2016}. The proof of existence and uniqueness of continuous MFG solutions is based on Picard's existence theorem \cite[Theorem 110C, and Appendix \ref{A0: Picards existence theorem}]{Butcher2016} and is well-established in the literature, e.g. \cite{Jourdain2007, Achdou2012, Carmona2013, CardaliaguetNotes}. Similarly, the existence and uniqueness of solutions for continuous-time, finite-state MFGs have also been thoroughly considered \cite{Gomes2013, Basna2014, Cecchin2020}. For the discrete MFG considered in Section \ref{sec: SIR}, the proof is provided by \citet{Doncel2019} and is based on Carathéodory’s Existence Theorem, which can be applied to discontinuous differential equations \cite[Appendix \ref{A1: Caratheodorys existence theorem}]{Biles1997}). As both proofs rely on the Lipschitz continuity of the functions $F$ and $G$, we can extend them to the data-driven case by assuming $\mu_\theta$ and $\sigma_\theta$ to be Lipschitz continuous as well.\footnote{This is a fair assumption to make as neural networks are generally formed by a composition of Lipschitz functions \cite{kidger2021}.} As long as the outputs of the neural networks are \emph{added} to the MFG dynamics, the resulting neural MFG consists of a sum of Lipschitz functions, which is again Lipschitz continuous.

\subsection{Implementation}
Even though it might be tempting to supply each agent with their own network to increase the diversity of strategies, this approach is not in line with the assumption of symmetry and drastically reduces the scalability of the method. Thus, we use a single neural network per MFG. More specifically, we use a multilayer perceptron with 3 to 8 hidden layers of 8 to 32 neurons. Even though the network seems very small, we stress that it's applied repeatedly during the differential equation solve, which increases its capacity. The network uses LipSwish activations to impose the Lipschitz condition \cite{kidger2021sde2} and is optimized with Adabelief \cite{adabelief}. The learning rate is $5 \times 10^{-4}$. The experimental framework has been implemented in Python. We use Optax, Diffrax, and Equinox to connect the methods to the JAX ecosystem \cite{kidger2021, kidger2021equinox, deepmind2020jax}. The proposed framework is lightweight, allowing all experiments to be performed on an Apple M1 Pro CPU with 16 GB of memory. 

\section{Toy Experiments
}
We demonstrate the capacity and flexibility of the neural MFGs by solving two example games. We start with the \textit{meeting arrival times} game introduced by \citet{Guéant2011}, a toy example considering a distribution of (unbounded) arrival times. With the \textit{El Farol Bar problem}, as posed by \citet{arthur1994inductive}, we discuss a more complex game that requires the estimation of probabilities. For each game, we extend the modelled drift with a neural network and describe how that influences the cost function associated with the MFG. 
\begin{figure}[b]
    \centering
    \includegraphics[width=0.45\textwidth]{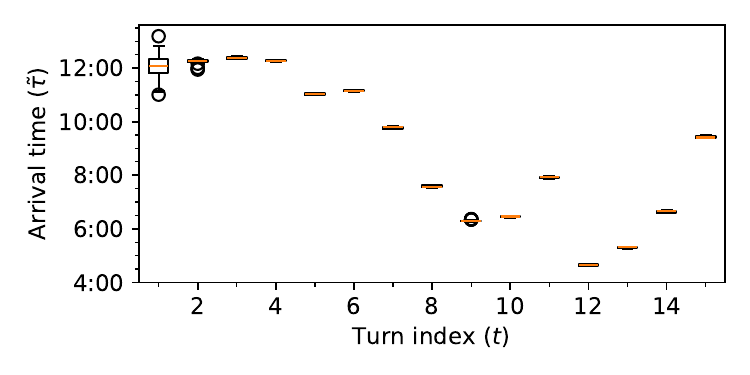}
    \caption{The distribution of meeting arrival times for the standard version of the \textit{meeting arrival times} game. The randomly initialized distribution of arrival times quickly converges to a narrow distribution centred at $\tilde{s}$. The value of $\tilde{s}$ is subject to the Brownian noise of the SDE.}
    \label{fig:mata}
\end{figure}

\begin{figure}[b]
    \centering
    \includegraphics[width=0.45\textwidth]{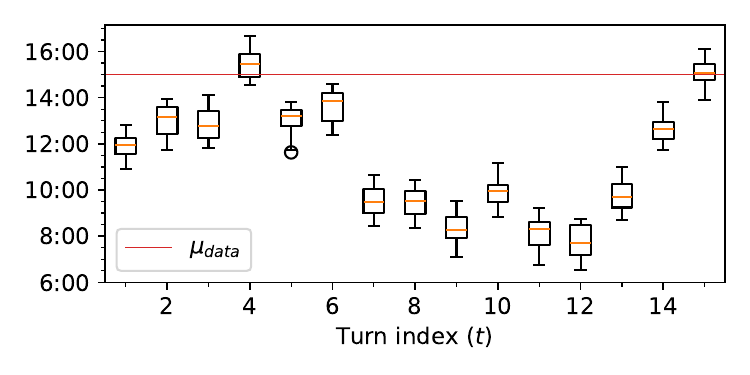}
    \caption{The distribution of meeting arrival times for the neural MFG version of the \textit{meeting arrival times} game. The distribution of \textit{meeting arrival times} of agents whose control is steered towards arriving at 15:00 in turn 15.\\}
    \label{fig:matb}
\end{figure}
\subsection{Meeting Arrival Times}
The \textit{meeting arrival times} game, introduced by \citet[p. 9]{Guéant2011}, illustrates a simple MFG application. The game considers a meeting that is scheduled at time $s$. Despite this planning, the true starting time of the meeting $\tilde{s}$ depends on the fraction of agents that have arrived. We assume that $\tilde{s} \geq s$, meaning that the meeting starts later if the quorum has not been reached at time $s$. Each agent $i$ has an intended arrival time $\tau^i$ and actual arrival time $\tilde{\tau}^i$. The agents' control $\tau$, but $\tilde{\tau}^i$ is subject to the uncertainty $\tilde{\epsilon}^i$ such that $\tilde{\tau}^i = \tau^i + \tilde{\epsilon}^i$. In our implementation of the game, $\tilde{\epsilon}^i$ is drawn from the Gaussian distribution $\mathcal{N}\left(0, 1\right)$. 

To build upon the formalism introduced in Section \ref{sec: MFG}, consider that the agents optimize their intended arrival time by minimizing their cost $J^i$. We consider the variant of the game that only has a terminal cost $G$, which is the sum of the components $g_{[1-3]}$ presented below. The notation $[\cdot]_+$ is shorthand for $\text{max}(\cdot, 0)$.\\
\noindent
    \begin{enumerate}[label=$g_\arabic*$, leftmargin=*]
    \item \textbf{The reputation effect} A cost associated with the lateness to the originally scheduled time $s$.
    $$g_1(s, \tilde{\tau}^i) = \left[\tilde{\tau}^i - s, 0\right]_+$$
    \item \textbf{A personal inconvenience} A cost associated with the lateness to the actual starting time $\tilde{s}$. 
    $$g_2(\tilde{s}, \tilde{\tau}^i) = [\tilde{\tau}^i - \tilde{s}, 0]_+$$
    \item \textbf{The waiting time} A cost associated with arriving earlier than the actual starting time $\tilde{s}$. 
    $$g_3(\tilde{s}, \tilde{\tau}^i) = [\tilde{s} - \tilde{\tau}^i , 0]_+$$
    \end{enumerate}

The behaviour of all other agents is summarized in the actual starting time $\tilde{s}$, which is the mean-field of this game that considers all agents' actions in a cumulative distribution of arrival times. To acknowledge the coupling between $\tilde{s}$ and the individual agents, consider that each agent bases their intended arrival $\tau^i$ time on $\tilde{s}$ and $\tilde{s}$ is influenced by the cumulative distribution of all actual arrival times $\tilde{\tau}^i$. In addition, \citet{Guéant2011} prove that $\tilde{s}$ exists, being a unique fixed point representing the game's equilibrium. 

We model the ``standard" version of the game with an SDE that differentiates over the turns $t$ (see Equation \ref{eq: dynamics}). Even though the game lacks a terminal condition, we treat it as a finite game by setting $t \in [1, 15]$. Figure \ref{fig:mata} shows an example outcome, where the distribution of arrival times quickly narrows down. This observation is in accord with the proof provided by \citet{Guéant2011}. The exact value of $\tilde{s}$ diffuses between turns due to the Brownian motion in the SDE. While the original game description does not include Brownian motion, it is included in this example to demonstrate the modelling capacity of the neural MFG. 
\begin{figure}[t]
    \centering
    \includegraphics[width=0.45\textwidth]{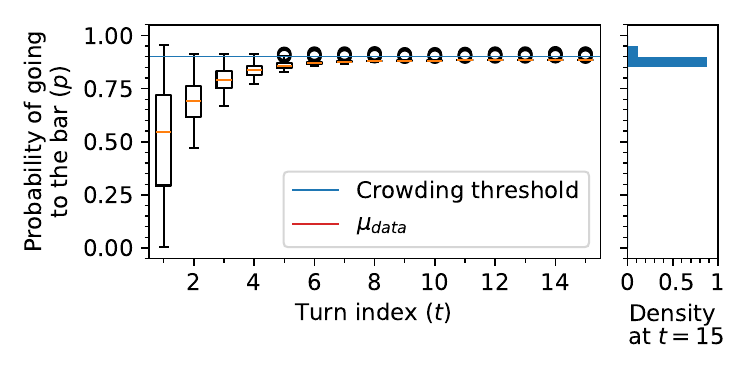}
    \caption{The distribution of probabilities of going to the bar for the standard version of the \textit{El Farol Bar problem}. In the standard version, the distribution of probabilities converges towards the crowding threshold of $c=0.9$ (blue). The density histograms show the normalized distributions at turn 15.}
    \label{fig:efbe}
\end{figure}
\subsubsection{Learning the Intended Arrival Times}
To apply the neural SDE, we create a data-driven challenge and aim to find the meeting time $s$ that aligns with a collection of observed arrival times. To this end, we convert the SDE into a neural SDE by adding the output of a feedforward neural network $f$. $f$ is trained to minimize the difference between the modelled arrival times at turn 15 and the ``true" observed arrival times, $\tilde{\tau}^*$. We train on 10 games. For each game, $\tilde{\tau}^*$ is generated by drawing $M$ samples from $N$ different Gaussians $\mathcal{N}\left(\mu_i, \sigma_i\right)$ for $i \in [1, 2, \dots, N]$. Each $\mu_i$ is sampled from $\mathcal{N}(15, 0.5)$, while $\sigma_i \sim \Gamma(2, 0.2)$. As a result, $\tilde{\tau}^*$ consists of $N \times M$ observations that form a noisy reflection of the distribution of intended arrival times around 15:00. The neural SDE models the interplay between the game rules and the observed data by minimizing the neural network loss, which can be seen as the evolving cost function $L$, alongside the agents' terminal cost $G$. The results are shown in Figure \ref{fig:matb}. The initial random distribution is centred around 12:00. Despite the small network size and the Brownian motion, the neural MFG converges around $s = \text{15:00}$ in turn 15.

\subsection{The El Farol Bar Problem}
Even though the meeting time arrival game provides a proof of concept, the neural SDE does not yield additional insights into the game dynamics. Therefore, we consider an implementation of the \textit{El Farol Bar problem}, a more complex example introduced by \citet{arthur1994inductive}. In this problem, a fixed population of agents can go to a bar. Each agent $i$ goes to the bar with a probability $p^i$, and all agents need to decide simultaneously whether they're going. The agents' utility depends on the bar's crowdedness. If the bar is too crowded, the agents who went there regret their decision and have less fun than the ones who stayed at home.

The cost function $J$ associated with the bar problem is comparable to that of the \textit{meeting arrival times} game. However, the new cost function is selective; not all components apply to all agents. This selectivity is presumed to lead to an interplay between the MFG dynamics and the neural network. The cost is parameterized by the agent's probability of going to the bar ($p^i$), the fraction of agents that went to the bar ($a$), and the crowding threshold ($c$). $J^i$ consists a terminal cost $G$ at state $T=15$, which is the sum of $g_{[1-3]}$ described below. 
\begin{enumerate}[label=$g_\arabic*$, leftmargin=*]
    \item \textbf{Missed a good evening} The cost associated with staying at home, while the bar was not too crowded. This cost only applies to agents who did not go to the bar.
    $$g_1(p^i, a, c) = \begin{cases}
    0 &  \text{if } a \geq c\\
    [c - p^i]_+ &  \text{if } a < c\\
    \end{cases}$$
    \item \textbf{Crowding penalty} The cost associated with going to a full bar. This cost only applies to the agents who went to the bar.
    $$g_2(p^i, a, c) = \begin{cases} 
     [p^i - c]_+ &  \text{if } a \geq c\\
     0 &  \text{if } a < c
                    \end{cases}$$
    \item  \textbf{Peer pressure} The cost inflicted due to not conforming with the rest of the agents. This cost does not depend on $C$ and applies to all agents.
    $$g_3(p^i, a) = \left(p^i - a \right)^2$$
\end{enumerate}

In contrast with the \textit{meeting arrival times} game, the value function of the \textit{El Farol Bar problem} does not include a Brownian motion. Instead, it is modelled as an SDE differentiating over $t \in [1, 15]$, where each turn presents a new opportunity to go to the bar. As the distribution of agents is summarized in the fraction of agents that went to the bar, $a$ can be considered the mean-field of the game. The agents reach a Nash equilibrium when the mean probability of going to the bar converges to the crowding threshold \cite{arthur1994inductive}. This behaviour is highlighted in Figure \ref{fig:efbe}. The figure shows the evolution of the distribution of agents, where $p$ goes from a uniform distribution on the interval $\left[0.0, 0.1\right\rangle$ at turn 1, to a narrow distribution centred around $c=0.9$ at turn 15. 

\subsubsection{Observed Attendance Rates and Emerging Strategies} Whereas the dynamics observed in Figure \ref{fig:efbe} purely emerge from the game's setup, we test whether a combination of the game rules with observed attendance rates leads to more informed strategies. To generate observations, we consider $N\times M$ agents. To create a single data point $x \in \mathbb{R}^{10}$, the agents' attendance rates are monitored for 10 subsequent turns. The attendance rate of one evening is generated by summing over $j \in [1, 2, \dots, N \times M]$ samples drawn from $Bern\left(p_j\right)$. Just like the arrival times data, the vector of probabilities is created by drawing $M$ samples from $N$ different Gaussians $\mathcal{N}\left(\mu_i, \sigma_i\right)$ for $i \in [1, 2,\dots, N]$. The hyperparameters are the same for both games. However, the agents' intentions $p_j$ are limited to the domain $\left[0, 1\right]$ to represent a probability. 

\begin{figure}[t]
    \centering
    \includegraphics[width=0.45\textwidth]{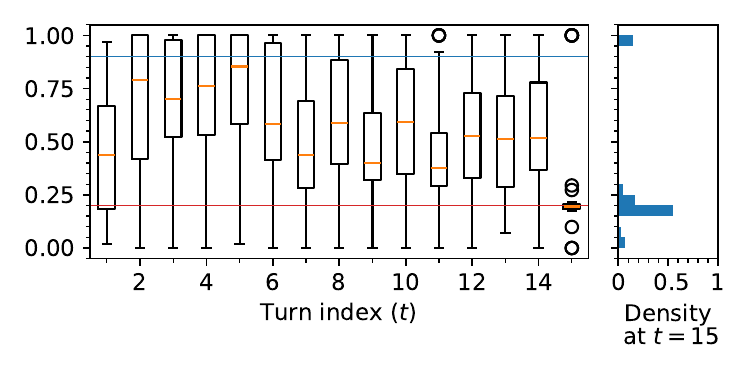}
    \caption{The distribution of probabilities of going to the bar for the neural MFG version of the \textit{El Farol Bar problem}. The neural MFG is trained to align with the observed attendance rates, where the mean $\mu_0=0.2$ is indicated in red. The density histograms show the normalized distributions at turn 15.}
    \label{fig:efbnMFG}
\end{figure}

To discover how the agents' probabilities of going to the bar change after training on 10 observations in $\mathbb{R}^{10}$, we deliberately diverge from the MFG setup by distributing the observed attendance rates around $\mu_0 = 0.2$ (Figure \ref{fig:efbnMFG}, red). Note that $\mu_0$ is significantly below the crowding threshold $c=0.9$ (Figure \ref{fig:efbnMFG}, blue). Even though the agents can only decide on a single output value, the neural MFG allows the agents to build more complex strategies than the standard MFG. Namely, the MFG-based agents in Figure \ref{fig:efbe} converge to an almost homogeneous strategy $p = c$. In contrast, the results from Figure \ref{fig:efbnMFG} are more widely distributed. Furthermore, the distribution ``oscillates" between turns. This oscillation results from the interplay between $j_1$, drawing the agents towards the crowding threshold, and the combination of the neural network contribution and $j_3$, pulling the agents back to $\mu_0$. Regardless of the oscillations, we observe agents with $p=1.0$, always going to the bar, independent of the other strategies. The agents from the last category leverage the fact that the majority of agents have a low probability of going to the bar. Because of this, they display emergent behaviour, and can always go to a bar that is never too crowded.

\section{The Simulation of Infectious Diseases}\label{sec: SIR}
In addition to the toy experiments described above, we use neural MFGs to simulate a real-world problem by considering a compartmental model for simulating the dynamics of COVID-19 in Japan. In particular, we analyze the susceptible, infected, and recovered (SIR) model of viral dynamics. SIR models from one of the simplest and most-studied models of virus propagation in a population \cite{Kermack1927}. While mathematical models can help to accurately predict the behaviour of infectious diseases, they often include assumptions on the relations between observations \cite{Marinov2022}. In addition, epidemic dynamics depend on population parameters and restrictive measures, features for which the effects may be challenging to quantify. Therefore, we extend the classical SIR model with a neural network to provide insights into the complex mechanics behind the spread of disease. 

The simulations presented in this work are based on the real-world observations of COVID-19 cases in Japan. The data reporting the number of confirmed infections, recoveries, and deaths, as well as any descriptions of restrictive measures, have been obtained from \citet{Edouard2020} and \citet{Hasell2020}. We extended the dataset to incorporate the number of administered vaccinations per day, based on the publications by the Japanese Ministry of Health, Labour and Welfare \cite{MOHJapan}. Whereas the original SIR model considers a finite population, we combine the works of \citet{Laguzet2015} and \citet{Doncel2022} to obtain an MFG definition. The governing equations of the resulting model are provided below. 
\subsection{A Mean-Field Game with a Finite Number of States} 
\begin{figure}[b]
    \centering
    \includegraphics[width=0.7\linewidth]{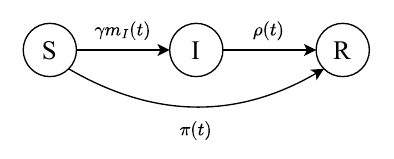}
    \caption{The state transition diagram of a single player in the SIR model. The figure describes the player's possible states: susceptible (S), infected (I), and recovered (R). The transitions between states depend on the transition parameters $\gamma$, $\rho$, and $\pi$, and the distribution of the population $\bm{m}$ at time $t$.}
    \label{fig:state_transition}
\end{figure}

We model the problem as an MFG with a finite number of states and consider a population of homogeneous players in continuous time $t \in [0, T]$. We consider three states in total, where each player can be susceptible ($S$), infected ($I$), or recovered ($R$). The mean-field of the game, $m(t)$, summarizes the fraction of players in each state. For ease of notation, we let the vector $m(t) = [m_S(t),\, m_I(t),\, m_R(t)]$ be the proportion of the population that is respectively in the states $S$, $I$, and $R$. The single-player dynamics are described by the following Markov process, which is summarized in Figure \ref{fig:state_transition}.
\begin{enumerate}
    \item Infected players ($I$) transfer the disease to susceptible ($S$) players with a transmission rate $\gamma$.
    \item Infected ($I$) players recover at a rate $\rho$, transitioning them to the recovered ($R$) state.
    \item Susceptible ($S$) players may choose to vaccinate at the rate $\pi(t)$, transferring them to the ``Recovered" ($R$) state. $\pi$ depends on $t$ to model a vaccination strategy that changes over time.  
\end{enumerate}
\begin{figure*}[b]
    \centering
    \includegraphics[width=0.90\textwidth]{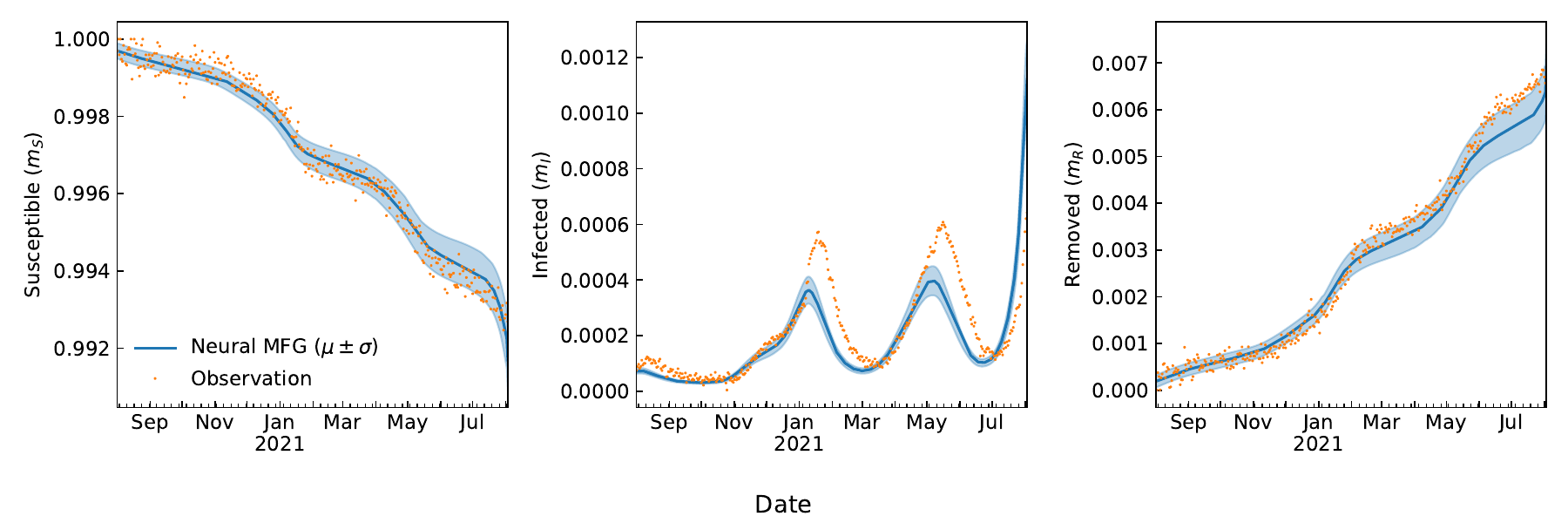}
    \caption{Comparison of the noisy, observed data (orange) and the predictions made with the neural MFG (blue) for the proportions of the Japanese population that are susceptible, infected, or removed. To visualize the predictions, we calculated the mean and standard deviation over the 100 noisy test samples. The results are normalized with respect to the total population to the domain [0, 1] and consider the period between August 1st, 2020, and August 3rd, 2021.}
    \label{fig: predictions sir neural mfg}
\end{figure*}
The player's state does not change once they have recovered from an infection or have been vaccinated. This observation simplifies most viral dynamics, as mutations typically enable (a version of) the virus to reinfect recovered or vaccinated players, although with a different rate $\sigma$. The exact SIR rates are unknown, and may even depend on the dominant virus variant at time $t$ \cite{Marinov2022}. As a consequence, SIR dynamics are ideally suited for data-driven inference.

\subsection{Modelling Viral Dynamics}The SIR model forms a Markov chain with nonlinear dynamics: the transition rates do not merely depend on the players' actions, but are influenced by their interaction with the population distribution $m$ as well \cite{Kolokoltsov2010, Doncel2016}. This introduces an explicit interaction between players that goes beyond the one facilitated through their cost function. The drift of the population is provided by the following Kolmogorov Equation, which is, in this case, a system of ordinary differential equations over a finite state space \cite{Doncel2022}:
\begin{equation}
\begin{cases}
    \dot m_S(t) = -\gamma(t) m_S(t)m_I(t) - \pi(t) m_S(t)\\
    \dot m_I(t) = \gamma(t) m_S(t)m_I(t) - \rho(t) m_I(t)\\
    \dot m_R(t) = \rho(t) m_I(t) + \pi(t) m_S(t)
    \end{cases}
\label{eq: population_dynamics}
\end{equation}

The solution for Equation \ref{eq: population_dynamics} is guaranteed to exist by Carathéodory’s Existence Theorem (Appendix \ref{A1: Caratheodorys existence theorem}) and can be found by solving the IVP \cite{Doncel2019}. We let $\mu_\theta: \mathbb{R}^d \to \mathbb{R}^r$ be the neural network extending the drift of the MFG, as proposed in Equation \ref{eq: neuralMFG dynamics}. In addition to directly influencing the population's state at time $t$, we also aim to learn the nonlinear dynamics of the SIR model by actively fitting the system's transition rates $\rho$, $\pi$, and $\gamma$. To do so, we let $\mu_\theta$ have $r=6$ output neurons, one for each state and rate. The resulting population drift is presented in Equation \ref{eq: neuralMFG population_dynamics}. In the equation, we omit the time dependence of the rates to simplify the notation.

\begin{equation}
    \begin{cases}
    \dot m_S(t) = -(\gamma + \mu_\theta^\gamma)  m_S(t)m_I(t) - (\pi + \mu_\theta^\pi) m_S(t) + \mu_\theta^S\\
    \dot m_I(t) = (\gamma + \mu_\theta^\gamma) m_S(t)m_I(t) - (\rho + \mu_\theta^\rho) m_I(t)  + \mu_\theta^I\\
    \dot m_R(t) = (\rho + \mu_\theta^\rho) m_I(t) + (\pi + \mu_\theta^\pi)m_S(t) +  \mu_\theta^R
    \end{cases}
\label{eq: neuralMFG population_dynamics}
\end{equation}
Here, the notation $\mu_\theta^x$ is used as shorthand for\\ $\mu_\theta^x(t, m_S(t), m_I(t), m_R(t), v(t))$, the neural network output corresponding to rate or state $x$ at time $t$. The vector $v$ denotes the auxiliary information that has been supplied to the neural network, which will be discussed in more detail in Section \ref{sec: aux}. We train the model on 200 epochs on 100 trajectories that are augmented with Gaussian noise (where $\varepsilon \sim \mathcal{N}(0, 0.05)$, scaled to the domain of the data). The model includes two neural networks, one for modelling the drift and one for modelling the diffusion. Each network has 8 hidden layers with 32 neurons each.

\subsubsection{A Data-Driven Evolving Cost}\label{sec:SIR_cost}
The cost $J$ associated with the drift from equation $\ref{eq: neuralMFG population_dynamics}$ has now become data-driven, where the neural network output $\mu_\theta$ partially describes the players' strategies. Because the observed data only includes measurements describing the dynamics of the population, we minimize a population-based cost $J$, of which the evolving loss $G$ is defined by the difference between the predicted, $\hat{m}^N_s$, and $m^N_s$ observed state of the population at time $s$. Due to the imbalance within the ratio susceptible, infected, and recovered players, we scale the loss by 1, $10^4$, and $10^3$ for the states $S$, $I$, and $R$, respectively. Here, we capitalize on the residuals of the infected state to accurately capture the quickly changing dynamics. We do not model a terminal loss, i.e., $G(x_T^i, m(T)) = 0$. The resulting loss is defined below

\begin{equation}
    J = \mathbb{E}\left[\int_{t_0}^T \frac{1}{2} L(\hat{m}^N(s), m^N(s))ds )\right].
\end{equation}

To provide the neural MFG with a warm start when optimizing the transition rates, we make an initial rate estimate. The details of this estimation are presented in Appendix \ref{appendix: experimental details}.
\subsubsection{Incorporating Auxiliary Observations}\label{sec: aux} 
Because the MFG drift as described in Equation \ref{eq: neuralMFG population_dynamics} is merely based on $m(t)$, it cannot incorporate any measures that may influence the state transition rates. For example, measures including school closings, lockdowns, and transport restrictions may negatively correlate with the virus's reproduction rate $\gamma$, whereas the rate may increase due to the 2020 Summer Olympics held in Tokyo from 23 July until 8 August 2021. In addition, the number of recovered citizens directly depends on the number of vaccinations administered. We incorporate these observations into the neural MFG via the vector $v$, which describes the status (0: not present, 1: moderate, 2: severe) of the COVID-19 restrictions at time $t$. In total, we model 7 different measures, leading to a total input size of $d=7 + 3 + 3=13$, combining $m(t)$, the transition rates, and measures.

\section{Experimental Setup and Results}
\subsection{Datasets}
We evaluate our framework on the publicly available COVID-19 pandemic datasets that have been published by \citet{Edouard2020, Guidotti2020} and \citet{Guidotti2022}. Rather than the full dataset, we focus on the disease dynamics reported in Japan, as these observations are reported daily, with fewer missing entries compared to other countries. We combine these data with the number of vaccinations administered, retrieved from \citet{MOHJapan}, to obtain a final dataset with the SIR dynamics, and 24 additional features describing the states. We refer to the task of modelling the viral dynamics from August 1st, 2020, and August 3rd, 2021 as the ``standard" dataset in Table \ref{tab:results}.  A visualization and description of the generation of the training data are provided in Appendix \ref{appendix: experimental details}.
\subsection{Baselines}
The neural MFGs combine continuous sequence prediction neural differential equations with mean-field dynamics. For this reason, we benchmark our approach against models that describe both categories. For the general prediction of continuous sequences, we include latent SDEs \cite{Li2020} ContiFormer \cite{Chen2023}, and Neural Laplace \cite{Holt2022}. In addition, we consider the model-free deep reinforcement learning algorithms Deep Average-network Fictitious Play (D-AFP) \cite{Lauriere2022}, and Deep Munchausen Online Mirror Descent (D-MOMD) \cite{Lauriere2022} to highlight RL-based approaches that are commonly used to model MFGs. The exact initialization of all baselines is presented in Appendix \ref{appendix: experimental details}. Table \ref{tab:results} presents the comparison across baselines, where our neural MFGs achieve a significant increase in performance over the benchmarks. 
\begin{table}[h]
    \centering
    \caption{Mean squared error (MSE) and mean absolute error (MAE) associated with the predictions on the COVID-19 SIR dynamics in Japan. Best-performing scores are highlighted in bold. }
    \resizebox{\columnwidth}{!}{
    \begin{tabular}{c | c c | c c}
    \toprule
       \multirow{2}{*}{\textbf{Method}} & \multicolumn{2}{c|}{\textbf{SIR} (standard)} & \multicolumn{2}{c}{\textbf{SIR} (extended)}\\
         & MSE & MAE & MSE & MAE \\
    \midrule
    Latent SDE     & $1.086 \times 10^{-3}$ & $2.324\times 10^{-2}$ & $3.185 \times 10^{-1}$ & $3.514 \times 10^{-1}$ \\
    ContiFormer    & $6.195 \times 10^{-1}$ & $8.393 \times 10^{-1}$ & $7.066 \times 10^{-1}$ & $8.406 \times 10^{-1}$ \\
    Neural Laplace & $8.850\times 10^{-2}$ & $1.994 \times 10^{-1}$ & $3.879$ & $1.591$ \\
    \midrule
    D-AFP          &$5.812 \times 10^{-1}$ &$6.219 \times 10^{-1}$ & - & - \\
    D-MOMD          & $8.698 \times 10^{-6}$& $2.005 \times 10^{-3}$  & - & - \\
    \midrule 
    Neural MFG     & $\bm{2.407 \times 10^{-7}}$ & $\bm{3.138 \times 10^{-4}}$ &$9.076 \times 10^{-6}$ & $2.181 \times 10^{-3}$ \\
    PDE MFG        & $1.979 \times 10^{-3}$& $2.439 \times 10^{-2}$ & $3.383 \times 10^{-1}$ & $3.360 \times 10^{-1}$\\
    Neural MFG (no aux)  & $2.677 \times 10^{-7}$& $3.304 \times 10^{-4}$ &$\bm{8.810 \times 10^{-6}}$ & $\bm{2.152 \times 10^{-3}}$ \\
    \bottomrule
    \end{tabular}}
    \label{tab:results}
\end{table}

\subsection{Ablation Studies}
\begin{figure*}[b]
    \centering
    \includegraphics[width=0.90\textwidth]{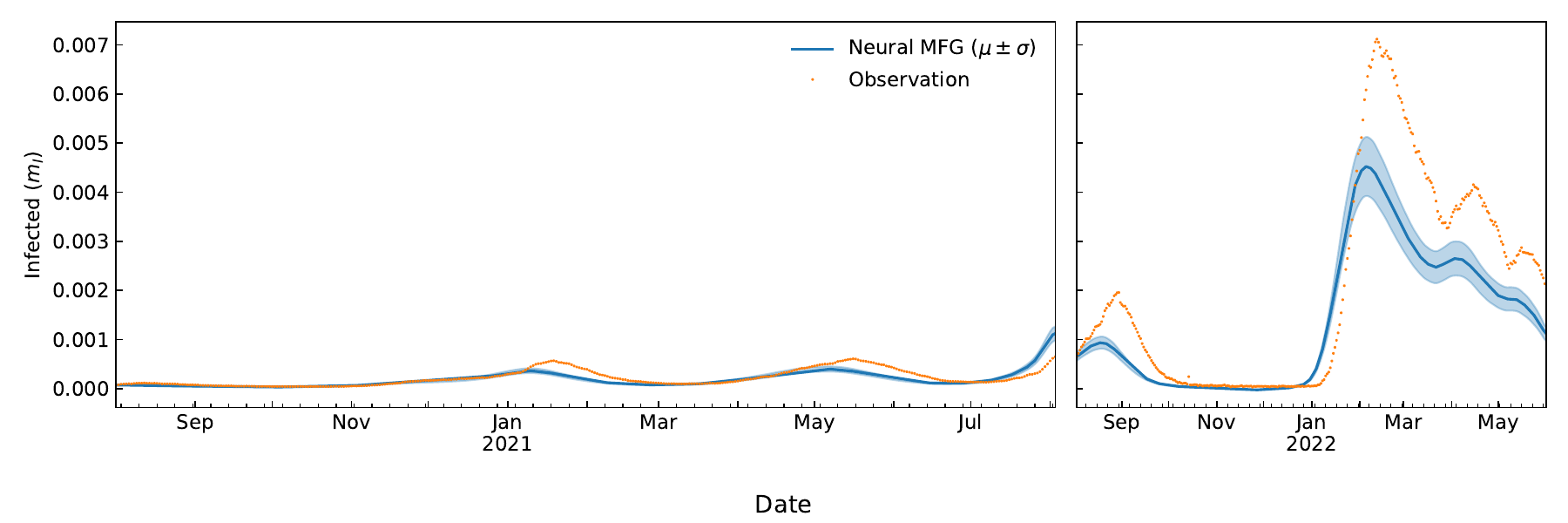}
    \caption{Out-of-sample predictions made with the Neural MFG (blue) for the proportion of the Japanese population that is infected. The neural MFG is trained on the period between August 1st, 2020, and August 3rd, 2021 (left pane), and tested on the period between August 1st, 2020, and June 1st, 2022 (right pane). To visualize the predictions, we calculated the mean and standard deviation over the 100 noisy test samples. The results are normalized with respect to the total population to the domain [0, 1].}
    \label{fig: oos predictions sir neural mfg}
\end{figure*}
\subsubsection{Long-term Extrapolation}
Whereas the main study focuses on learning the in-distribution dynamics, we also investigate the models' ability to extrapolate the epidemic dynamics across longer periods of time. To this end, we train the models on the SIR dynamics from August 1st, 2020, to August 3rd, 2021. Next, we use the same model to predict the viral dynamics between August 3rd, 2021, and June 1st, 2022. This task is referred to as the ``extended" dataset in Table \ref{tab:results}, and the results presented are merely calculated w.r.t. out-of-sample observations. We show a qualitative example of the out-of-sample predictions using the neural MFG in Figure \ref{fig: oos predictions sir neural mfg}. Here, we observe a massive shift in the viral dynamics around February 2022, picked up by the neural MFG: although the model does not fully capture this peak, it still manages to model the shape of the infections accurately. This indicates that the viral dynamics have changed compared to the training period. This change in dynamics may have been fueled by the dominance of the new, more contagious SARS-CoV-2 variant in Japan. For example, the sudden increase in infections correlates with the increase in dominance of Omicron BA.2 in Tokyo and the rest of Japan \cite{Takeuchi2025, Omicron2022}.

\subsubsection{Modelling the Mean-field Dynamics}
\begin{figure*}[t]
    \centering
    \includegraphics[width=\textwidth]{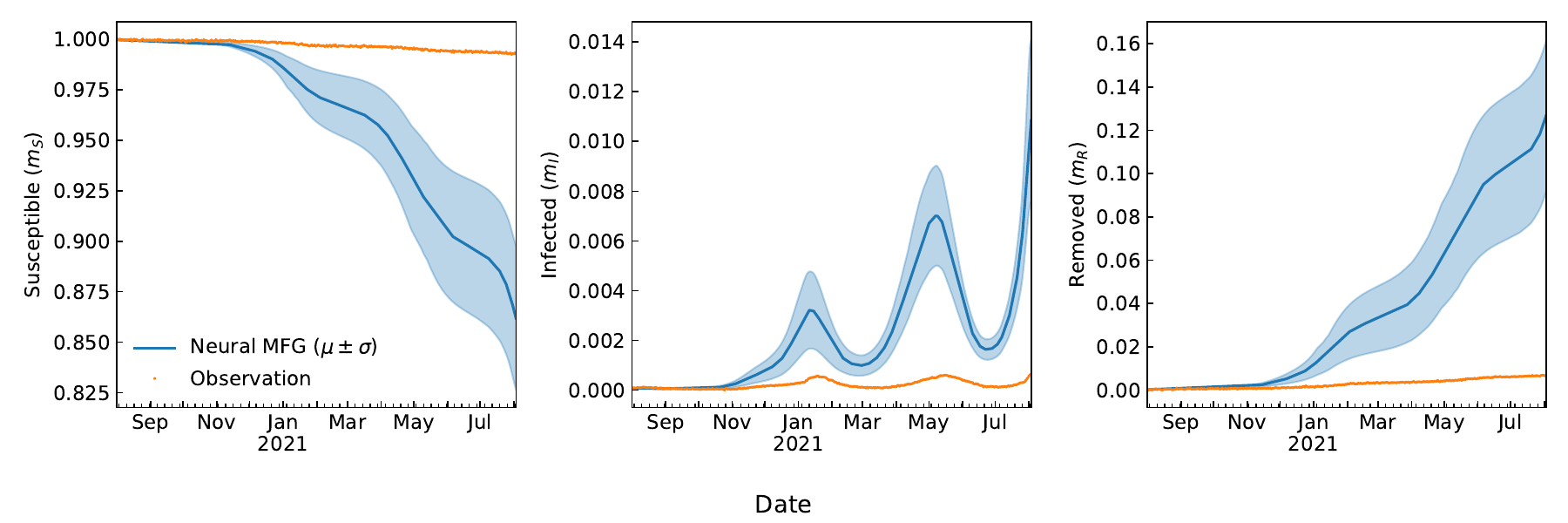}
    \caption{Predictions made with the PDE MFG (blue) with a predefined, deterministic drift based on Equation \ref{eq: population_dynamics} and a neural diffusion. To visualize the predictions, we calculated the mean and standard deviation over the 100 noisy test samples. The results are normalized with respect to the total population to the domain [0, 1] and consider the period between August 1st, 2020, and August 3rd, 2021.}
    \label{fig: predictions sir mfg}
\end{figure*}

The predictions made with the neural MFG, which are shown in Figure \ref{fig: predictions sir neural mfg}, capture the peaks in infections around January, May, and August 2021 (middle panel). While the model fit is not fully accurate, it is interesting to learn how a similar model without a neural drift component performs on the same task. To this end, we modelled the SIR dynamics using predetermined, deterministic drift based on Equation \ref{eq: population_dynamics} and a neural diffusion with the same parameterization as the diffusion in the neural MFG. The resulting model, referred to as PDE MFG, performs less well than the neural MFG (Figure \ref{fig: predictions sir mfg}). Even though the model succeeds in capturing the shape of the viral dynamics, the scale is incorrect. In addition, the model tends to overestimate the peak of infections observed around May 2021, suggesting that the addition of $v$ enables the neural MFG to more accurately predict the viral dynamics. 
\subsubsection{The Importance of Auxiliary Observations}
To assess the information conveyed by the inclusion of the auxiliary variables, we train the neural MFG without this additional information. We included the performance of the resulting method in Table \ref{tab:results} under ``neural MFG (no aux)". The inclusion of auxiliary observations leads to a small increase in performance, highlighting the informativeness of the additional features. Interestingly, this observation only holds within-sample. For the out-of-sample predictions associated with the extended dataset, the neural MFG without auxiliary observations outperforms its more informed counterpart.

\section{Discussion}
Neural MFGs form a powerful tool for influencing the Nash equilibria of a game. Our experimental results highlight the model's strengths: the results found with \textit{meeting arrival times} game underline its ability to learn abstractions from data, even under noisy circumstances. In addition, the analyses of the \textit{El Farol Bar problem} exemplify strategies that emerge from the interaction between the data-driven dynamics and the MFG mechanics. Lastly, the experiments with the \emph{SIR model} have shown that the neural MFG can learn elaborate data-driven strategies based on auxiliary observations. This alleviates the challenge of manually designing an accurate model of the framework at hand. \\
\textbf{Limitations} The addition of the neural network introduces a ``black box" into the game dynamics. In particular, the inclusion of the neural network reduces the game's general interpretability by making the agents' decisions more opaque. Furthermore, the formulation of the model is limited by the symmetry assumption of the MFG. Because of the large number of players involved, the neural network has to be shared between agents. While this does not imply that the agents have identical beliefs, they are still linked through the neural network parameters. 
\section{Conclusion}
Mean-field game theory is used to efficiently model games with a very large to infinite population of agents. The agents' Nash equilibria are commonly found via reinforcement learning or by solving the system of PDEs associated with the MFG. Both methods have their advantages, the first being model-free and the second numerically exact. To get the best of both worlds, we extend the PDE formulation of the MFG with a neural network. 
The resulting neural MFG exerts a data-driven influence on the game's Nash equilibria, deriving strategies that are in line with observed game outcomes. In addition, it can be trained to recover structure from the game's past dynamics, learning more complex strategies without using extra data. Leveraging those properties, the model can fill in dynamics that are unaccounted for in the original MFG specification.
Besides their ability to incorporate observations, neural MFGs are a numerically exact and lightweight alternative to (deep) RL-based methods for solving MFGs. To match the RL-based methods in their flexibility, the neural MFG can be made completely model-free by fully omitting the predefined partial derivatives.
Lastly, their dependency on automatic differentiation makes the neural MFGs more objective than approaches that solve PDEs using finite differences \cite{chen2018neural, SU1997}, which may prevent the loss of existence or uniqueness of solutions associated with solving MFGs \cite{RL-basedNNs}.
We translate the theoretical advantages stated above into an impactful application by successfully modelling the outbreak of an infectious disease. Hereby, our work helps bridge the gap between mathematical modelling and real-world applications, outperforming all baselines considered. Even though we highlight a biological application, we expect the model to find applications in any system associated with non-atomic, anonymous players and observations. Such systems are, e.g., also commonly found in markets and economics \cite{Schmeidler1973EquilibriumPO}. 

\section*{Acknowledgements}
Tal Kachman and Anna C.M. Thöni acknowledge funding from the National Growth Fund project “Big Chemistry” (1420578) and the European Laboratory for Learning and Intelligent Systems (ELLIS). The authors also thank Mathieu Laurière for the insightful discussions.

\bibliography{sn-bibliography}

\newpage
\appendix
\onecolumn
\section{Theorems of Existence and Uniqueness}
\subsection{Picard's Existence Theorem}\label{A0: Picards existence theorem}
The Picard-Lindelöf Theorem, also known as Picard's Existence Theorem, gives a set of conditions under which an initial value problem has a unique solution \cite[Theorem 110C]{Butcher2016}. 

\begin{theorem} (Picard's Existence Theorem)
Let $f: [0, T] \times \mathbb{R}^d \to \mathbb{R^d}$ be continuous in t and Lipschitz continuous in y, and let $y_0 \in \mathbb{R}^d$. Then there exists a unique differentiable $y: [0, T] \to \mathbb{R}^d$ that satisfies

$$y(0) = y_0 \; \; \; \frac{dy}{dt}(t) = f(t, y(t)).$$
\label{theorem:picard}
\end{theorem}
\subsection{Carathéodory’s Existence Theorem}\label{A1: Caratheodorys existence theorem}
The Carathéodory’s Existence Theorem is a fundamental theorem that guarantees the existence of solutions to certain initial value problems that include discontinuous differential equations \cite{Biles1997}. 
\begin{theorem} (Carathéodory’s Existence Theorem)
\label{theorem:Caratheodory}
Consider the initial value problem
$$y(t_0) = y_0 \; \; \; \frac{dy}{dt}(t) = f(t, y(t)).$$
Furthermore, let $f: [t_0, T] \times \mathbb{R}^d \to \mathbb{R^d}$ be measurable in $t$ for each fixed $y$, continuous in $y$ for each fixed $t$, and let $y_0 \in \mathbb{R}^d$. $f$ is defined on the rectangular domain $R= \{(t, y)| \; |t-t_0| \leq a, |y-y_0| \leq b\}$. In addition, let there be a Lebesgue-integrable function $m: [t_0 -a , t_0 + a] \to [0, \infty \rangle$ such that $|f(t, y(t)| \leq m(t)$ for all $t, y \in R$. Then the differential equation has a solution in the extended sense in the neighbourhood of the initial condition $y_0$.
\end{theorem}

\subsubsection{Continuous Time and Space}
For neural MFGs that consider a continuous time and space, we show that the sum of two systems of differential equations, each having a unique solution and being Lipschitz continuous, also has a unique solution. 
\begin{theorem}
    Let $f_1: [0, T] \times \mathbb{R}^d \to \mathbb{R^d}$ and $f_2: [0, T] \times \mathbb{R}^d \to \mathbb{R^d}$ be continuous in t and Lipschitz continuous in x and y, where $y_0 \in \mathbb{R}^d$ and $x_0 \in \mathbb{R}^d$. Consider the following systems of differential equations
    \begin{align*}
        \frac{dy}{dt}(t) &= f_1(t, y(t))\\
        \frac{dx}{dt}(t) &= f_2(t, x(t))\\
    \end{align*}
with initial conditions $y(t_0)=y_0$ and $x(t_0)=x_0$. If each system has a unique solution, then the system 

\begin{equation*}
    \frac{dz}{dt}(t) = f_1(t, z(t)) + f_2(t, z(t))
\end{equation*}
\noindent
with initial condition $z(t_0) = z_0$ also has a unique solution.
\end{theorem}

\begin{proof}
Let $f(t, z(t)):= f_1(t, z(t)) + f_2(t, z(t))$. Because both $f_1$ and $f_2$ are Lipschitz continuous in $z$, there exist constants $L_1>0$ and $L_2>0$ such that for all $z_1$, $z_2 \in \mathbb{R}^d$ it holds that 
\begin{align*}
    ||f_1(t, z_1(t)) - f_1(t, z_2(t))|| &\leq L_1 ||z_1(t) - z_2(t)||\\
    ||f_2(t, z_1(t)) - f_2(t, z_2(t))|| &\leq L_2 ||z_1(t) - z_2(t)||
\end{align*}
Where the notation $||\;||$ denotes vector norm. Substituting the definition of $f(t, z(t)):=f_1(t, z(t))+f_2(t, z(t))$ and expanding the difference for an arbitrary $z_1$ and $z_2$, we obtain
\begin{equation*}
    ||f(t, z_1(t)) - f(t, z_2(t))|| = ||f_1(t, z_1(t)) + f_2(t, z_1(t)) - f_1(t, z_2(t)) - f_2(t, z_2(t))||.
\end{equation*}
From the triangle inequality, it follows that
\begin{align*}
    ||f(t, z_1(t)) - f(t, z_2(t))|| &= ||f_1(t, z_1(t)) + f_2(t, z_1(t)) - f_1(t, z_2(t)) - f_2(t, z_2(t))||\\
    &\leq ||f_1(t, z_1(t)) - f_1(t, z_2(t))|| + ||f_2(t, z_1(t)) - f_2(t, z_2(t))||\\
    &\leq L_1 ||z_1(t) - z_2(t)|| +  L_2 ||z_(t)1 - z_2(t)|| = (L_1 + L_2)||z_1(t) - z_2(t)||
\end{align*}
Therefore, $f$ is Lipschitz continuous with respect to $z$ with Lipschitz constant $L:=L_1+L_2$. Then, using Picard's Existence Theorem (Appendix \ref{A0: Picards existence theorem}), we can conclude that the initial value problem

$$z(t_0) = z_0 \; \; \; \frac{dz}{dt}(t) = f(t, z(t))$$
\noindent
has a unique solution.

\end{proof}
\subsubsection{Continuous Time and Discrete Space}
Similarly, we can provide the proof of existence and solutions of mean-field games with discrete states based on Carathéodory’s Existence Theorem (Appendix \ref{A1: Caratheodorys existence theorem}):

\begin{theorem}
Consider the initial value problem 

$$z(t_0) = z_0 \; \; \; \frac{dz}{dt}(t) = f_1(t, z(t)) + f_2(t, z(t))$$

where $f_1: [t_0, T] \times \mathbb{R}^d \to \mathbb{R}$ and $f_2: [t_0, T] \times \mathbb{R}^d \to \mathbb{R}$ be measurable in $t$ for fixed $z$, and continuous in z for each fixed t. Both $f_1$ and $f_2$ are defined on the rectangular domain $R= \{(t, y)| \; ||t-t_0|| \leq a, ||y-y_0|| \leq b\}$. In addition, let there be two Lebesgue-integrable function $m_1: [t_0 -a , t_0 + a] \to [0, \infty \rangle$ and $m_2: [t_0 -a , t_0 + a] \to [0, \infty \rangle$ such that $||f_1(t, z(t)|| \leq m_1(t)$ and $||f_2(t, z(t)|| \leq m_2(t)$ for all $t, y \in R$. In that case, there exists a solution to the initial value problem in the extended sense in the neighbourhood of the initial condition $z_0$.
\end{theorem}
\begin{proof}
The proof is split into the three conditions required to apply Carathéodory's Existence Theorem:
\begin{enumerate}
    \item \textbf{Measurability in t}. Because both functions $f_1$ and $f_2$ are measurable in $t$ for each fixed $z$, then $f(t, z(t)):=f_1(t, z(t)) + f_2(t, z(t))$, being the sum of two measurable functions, is also measurable. This requirement is met if $f_1$ and $f_2$ are Lipschitz with respect to their second argument, which makes them continuous and measurable in $z$. 
    \item \textbf{Continuity in z}.
    Because both functions $f_1$ and $f_2$ are continuous in $z$ for each fixed $t$, then $f(t, z(t)):=f_1(t, z(t)) + f_2(t, z(t))$, being the sum of two continuous functions, is also continuous.  This requirement is met if $f_1$ and $f_2$ are Lipschitz with respect to their second argument, which makes them continuous and measurable in $z$. 
    \item \textbf{Local integrability bound}. Because $f_1$ and $f_2$ are bounded by $m_1$ and $m_2$ respectively, it holds by the triangle inequality that
\begin{align*}
    ||f(t, z(t))|| &= ||f_1(t, z(t) + f_2(t, z(t))|| \\
    & \leq ||f_1(t, z(t))|| + ||f_2(t, z(t))|| \\
    & \leq m_1(t) + m_2(t).
\end{align*}
\noindent
If we define $m(t) := m_1(t) + m(2)$, then we can state that $||f(t, z(t))|| \leq m(t)$. Having satisfied the three conditions, it follows from Carathéodory's Existence Theorem that the initial value problem has a solution in the neighbourhood of the initial condition.
\end{enumerate}
\end{proof}
\section{Experimental Details}\label{appendix: experimental details}
\subsection{Generating the Training Data}\label{training_data}
To learn the viral dynamics, we treat the reported number of susceptible, infected and recovered people as the ground truth data labels $\bm{\mu}(t) = [m_S(t),\, m_I(t),\, m_R(t)]$, and sample $N=100$ noisy training observations from a Gaussian distribution $\mathcal{N}(\bm{\mu}(t), \bm{\sigma})$, where $\bm{\sigma} = [2.5 \times 10^{-4}, 5 \times 10^{-2}, 5 \times 10^{-2}]$. We use a different noise scale per state to account for the state imbalances.

\begin{figure}[H]
    \centering
    \includegraphics[width=\textwidth]{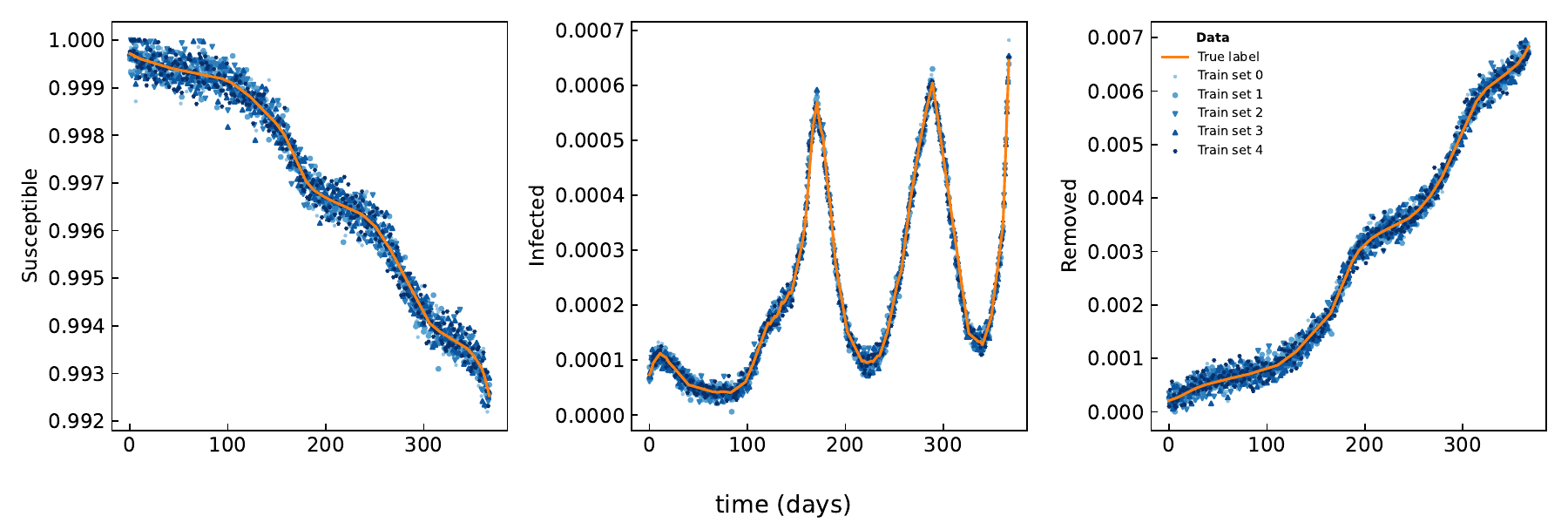}
    \caption{Ground truth observations (orange line) and five example data points (train set 0-4, scattered markers).}
    \label{fig:placeholder}
\end{figure}

\subsection{Estimating the Transition Rates in a SIR Model}\label{A: rate_estimate}
We make an initial estimate of the SIR transition rates $\kappa$ to provide the MFG-based models with a warm start for optimizing the SIR dynamics. The initial estimate is based on a minimization of the mean squared error (MSE) between the noiseless data $y$ and the predictions $\hat{y}$ associated with an integration that is parameterized with the current estimate $\hat{\kappa}$. To avoid smoothing the rates too much, we estimate daily rates based on a rolling, 28-day window. We use the Nelder-Mead simplex algorithm to solve the optimization problem \cite{Gao2012}. Because the data can become very stiff if the wrong rates $\hat{\kappa}$ are selected, the numerical computation of the optimization derivative is rather unstable. Being a simplex algorithm, Nelder-Mead provides sufficient estimates for a warm start.

\subsection{Baseline Initialization}\label{A: baseline initialization}
All baseline models are trained for 200 epochs. For ContiFormer \cite{Chen2023}, Neural Laplace \cite{Holt2022}, Deep Average-network Fictitious Play (D-AFP), and Deep Munchausen Online Mirror Descent (D-MOMD) \cite{Lauriere2022}, we use the default parameterization. The Latent SDE is parameterized with a multilayer perceptron with 8 hidden layers with 32 neurons each 32. For D-AFP and D-MOMD, we translated the SIR MFG to OpenSpiel \cite{lanctot2020}. Here, we incorporated the Covid-19 data in the game's rewards, combining the rewards proposed by \citet{Cui2021} with a scaled MSE loss as presented in Section \ref{sec:SIR_cost}.

\subsection{Ablation on the Network Size}
We present a grid-based ablation study on the neural network size below. From Table \ref{tab:network_size}, it becomes clear that networks with 16 layers tend to overfit. The layer width forms an important aspect of the modelling capacity, where it seems that networks with narrow layers ($<$16 neurons) do not perform as well as networks with broader ($\geq$ 16 neurons) layers.
\begin{table}[h]
    \centering
    \caption{Overview of the validation loss for various neural network sizes. }
    \begin{tabular}{l l l}
    \toprule 
     \textbf{Network Depth} & \textbf{Layer Width} & \textbf{Validation Loss}	   \\
     \hline
    2	&4	&2.60		            \\
    2	&8	&0.921	                \\
    2	&16	&0.532		            \\
    2	&32	&0.478	                \\
    4	&4	&5.20		            \\
    4	&8	&0.7320		            \\
    4	&16	&0.5758		            \\
    4	&32	&0.5163		            \\
    8	&4	&37.35	                \\
    8	&8	&71.11		            \\
    8	&16	&0.5948	                \\
    8	&32	&0.5990	                \\
    16	&4	&8.208	                \\
    16	&8	&0.7355		            \\
    16	&16	&0.7004	                \\
    16	&32	&0.6128		            \\
    \bottomrule 
    \end{tabular}
    \label{tab:network_size}
\end{table}

\end{document}